\documentclass{article}

\usepackage{graphicx}

\usepackage{amssymb,amsmath,epsfig}
\usepackage{algorithm}
\usepackage{algorithmic}
\usepackage{mathtools}
\usepackage{lscape}
\usepackage{mathrsfs}
\usepackage{bm}
\begin{document}

% Keywords command
\providecommand{\keywords}[1]
{
  \small	
  \textbf{\textit{Keywords---}} #1
}

\title{\bf Multivariate Density Estimation with Deep Neural Mixture Models}


%\titlerunning{Neural Mixture Models}        % if too long for running head

\author{Edmondo Trentin\\
DIISM (Universit\`a di Siena, Italy)\\
%Via Roma, 56 - 53100 Siena (Italy)\\
 \tt{trentin@dii.unisi.it}           %  \\
}

%\authorrunning{E. Trentin} % if too long for running head

%\date{Dec 5, 2020}
\date{}

\maketitle

\begin{abstract}
Albeit worryingly underrated in the recent literature on machine learning in general (and, on deep learning in particular), multivariate density estimation is a fundamental task in many applications, at least implicitly, and still an open issue. With a few exceptions, deep neural networks (DNNs) have seldom been applied to density estimation, mostly due to the unsupervised nature of the estimation task, and (especially) due to the need for constrained training algorithms that ended up realizing proper probabilistic models that satisfy Kolmogorov's axioms. Moreover, in spite of the well-known improvement in terms of modeling capabilities yielded by mixture models over plain single-density statistical estimators, no proper mixtures of multivariate DNN-based component densities have been investigated so far. The paper fills this gap by extending our previous work on Neural Mixture Densities (NMMs) to multivariate DNN mixtures. A maximum-likelihood (ML) algorithm for estimating Deep NMMs (DNMMs) is handed out, which satisfies numerically a combination of hard and soft constraints aimed at ensuring satisfaction of Kolmogorov's axioms. The class of probability density functions that can be modeled to any degree of precision via DNMMs is formally defined. A procedure for the automatic selection of the DNMM architecture, as well as of the hyperparameters for its ML training algorithm, is presented (exploiting the probabilistic nature of the DNMM). Experimental results on univariate and multivariate data are reported on, corroborating the effectiveness of the approach and its superiority to the most popular statistical estimation techniques. 
\end{abstract}


\section{Introduction}
\label{sec:intro}

Let us consider an unlabeled training set ${\cal T} = \{{\bf x}_1, \ldots, {\bf x}_n\}$ of $n$ independent random vectors (i.e., patterns) in a $d$-dimensional feature space, say $\mathbb{R}^d$. The patterns are assumed to be identically distributed according to an unknown probability density function (pdf) $p({\bf x})$. Density estimation consists in finding a model for $p({\bf x})$ based on ${\cal T}$. This requires exploiting the statistical knowledge on $p({\bf x})$ implicitly underlying the very data sample ${\cal T}$. To this aim, a suitable algorithm is applied, either rooted in statistics or in machine learning. The algorithm is expected to come up with a model of $p({\bf x})$ that fits the data well, in compliance with some well-defined statistical criterion  (the maximum likelihood criterion \cite{DudHar73} being possibly the most popular). The present paper proposes and investigates mixtures of multivariate component densities realized via deep neural networks (DNNs) for density estimation. It extends to multivariate mixture densities our previous work on univariate pdfs, presented as a workshop communication in \cite{Trentin-NMM-ANNPR18}. The reader is referred to that paper for a representative list of up-to-date applications where (implicitly or explicitly) density estimation is fundamental and still an open issue.

\subsection{Intrinsic difficulties of density estimation via machine learning} 

As pointed out by Vapnik \cite{Vapnik95}, density estimation is an intrinsically difficult problem, and it is still open nowadays. This latter fact is mostly due to the shortcomings of established statistical approaches, either parametric or non-parametric (the reader is referred to \cite{TrentinLC18} for a list of the major drawbacks of the statistical techniques), and by the technical difficulties that arise from attempting to use artificial neural networks (ANNs) or machine learning for pdf estimation. Such difficulties stem from: (1) the unsupervised nature of the learning task, (2) the numerical instability problems entailed by pdfs, whose codomains may span the interval  $[0, +\infty)$, and (3) the requirement for the resulting model to respect the axioms of probability. Furthermore, the use of maximum-likelihood (ML) training in ANNs tends to result in the ``divergence problem'', observed first in the realm of hybrid ANN/hidden Markov models \cite{Trentin15_ML}. It consists in the progressive divergence of the value of the ANN connection weights as ML training proceeds, resulting in an unbounded growth of the integral of the pseudo-pdf computed by the ANN. The problem does not affect radial basis functions (RBF) networks whose hidden-to-output weights were constrained to be positive and to sum to one, as in the RBF/echo state machine for sequences proposed in \cite{TrentinSS15_ESN}, or in the RBF/graph neural network presented in \cite{BonRigTre2018} for the estimation of generalized random graphs. Unfortunately, the use of RBFs in the latter contexts is justified by its allowing for a proper algorithmic hybridization with models devised specifically for sequence/structure processing, but using RBFs as a stand-alone paradigm for density estimation is of neglectable practical interest, since they end up realizing plain Gaussian mixture models (GMM) estimated via ML.

\subsection{Earlier approaches to ANN-based density estimation}
  
In spite of these difficulties, several approaches to pdf estimation via ANNs are found in the literature \cite{TrentinFreno09}. A critical survey of the literature is found in \cite{TrentinLC18}. Most of these approaches suffer from some limitations. First, a ML technique is presented in \cite{ModhaFainman94} where the ``integral equals 1'' requirement is satisfied numerically by dividing the output of a multilayer Perceptron (MLP)  by the numerical integral of the function the MLP computes. No algorithms for computing the numerical integral over high-dimensional spaces are handed out in \cite{ModhaFainman94}. Nonetheless, this approach is related to the technique presented in this paper, insofar that ML will be exploited herein. Differently from \cite{ModhaFainman94}, a multi-dimensional ad-hoc numeric integration method will be used in the following, jointly with hard constraints, over a mixture of DNNs.

Instead of estimating the pdf directly, techniques have been proposed that focus on the (theoretically equivalent) estimation of the corresponding cumulative distribution function (cdf) \cite{Vapnik00supportvector,IsmailAtaiya2002den}. Supervised training is used , e.g. the backpropagation algorithm (BP), relying on the empirical cdf of the data for generating the target outputs required in supervised training.  After training the MLP model $\phi(\cdot)$ of the cdf, the pdf can be retrieved by taking derivatives of $\phi(\cdot)$. Unfortunately, there are drawbacks to these approaches, as well (see \cite{TrentinLC18}). In particular, a good approximation of the cdf does not entail a good estimate of its derivative. Negative values of $\frac{\partial \phi(x)}{\partial x}$ may even occur, since a linear combination of logistics is not necessarily monotonically increasing. Finally, these algorithms apply naturally to univariate cases, but their application to  multivariate scenarios is way trickier (to say the least).

The idea of exploiting synthetically-generated target outputs was applied to DNN-based pdf estimation in \cite{TrentinLC18} and \cite{Trentin_Soft_NEPL}.  The former relies on a unbiased variant of the Parzen-window (PW) density estimation technique \cite{DudHar73} for labeling the training set for a DNN, known as Parzen neural network. Like in the traditional $k_n$-Nearest Neighbor ($k_n$-NN) statistical technique \cite{DudHar73}, the resulting model does not satisfy the axioms of probability. To the contrary, the algorithm for density estimation via feed-forward DNNs presented in \cite{Trentin_Soft_NEPL} uses a modified loss function, optimized via gradient descent. Such a criterion function combines two terms: the loss between the network output and a synthetically-generated non-parametric estimate of the pdf at hand evaluated over the corresponding input pattern, and a loss between the integral of the function realized by the MLP and the unity value (that is the training ``target'' for the integral of the DNN). Efficient integration methods based on Markov chain Monte Carlo with importance sampling are used to compute the integral of the DNN and its derivatives w.r.t. the DNN parameters within the gradient-descent scheme used for the optimization. The asymptotic convergence of the resulting technique to the correct solution was formally proven in \cite{TrentinMathematics2020}. The ideas behind such integration methods are exploited in this paper, as well.

\subsection{Mixture densities}

A generalization of plain pdf models stems from the adoption of mixture densities, where the unknown pdf is rather modeled in terms of a combination of any number of component densities \cite{DudHar73}. GMMs are the most popular instance of mixture densities \cite{Bishop:2006:PRM:1162264}. Mixture densities were originally intended as real-life extensions of the single-pdf parametric model (e.g., one Gaussian may not be capable to explain the data distribution but $K$ Gaussian pdfs might as well be). Yet, there is much more than this behind the notion of mixture density. In fact, in mixture densities the different component pdfs are specialized to explain distinct latent phenomena (e.g., stochastic processes) that underlie the overall data generation process, each such phenomenon having different likelihood of occurrence w.r.t. others at diverse regions of the feature space. This suites particularly those situations where the statistical population under analysis is composed of several sub-populations, each having different distribution. Relevant examples are the following:

\begin{enumerate}

\item the {\em statistical study of heterogeneity} in meta-analysis \cite{metanalysis}, where samples drawn from disjoint populations (e.g., adults and children, male and female subjects, etc.) are collectively collected and have to be analyzed as a whole;

\item the modeling of {\em unsupervised} or {\em partially-supervised} \cite{SchwenkerT14a_Survey} data samples in statistical pattern recognition \cite{DudHar73}, where each sub-population corresponds to a class or category;

\item the {\em distribution of financial returns} on the stock market depending on latent phenomena such as a political crisis or a war \cite{finance-crisis};

\item the {\em assessment of projectile accuracy} in the military science of ballistics when shots at the same target come from multiple locations and/or from different munition types \cite{CIS-106239}.

\end{enumerate}

As it happens, the sub-populations in a mixture are unlikely to be distributed individually according to simple (e.g., Gaussian) pdfs. Consequently, parametric models  (like GMMs) are generally not a very good fit. In fact, let $\xi_1, \ldots, \xi_K$ be $K$ disjoint states of nature (the outcomes of a discrete, latent random variable $\Xi$, each outcome corresponding to a specific sub-population), and let $p({\bf x}|\xi_i)$ be the pdf that explains the distribution of the random observations ${\bf x}$ given the $i$-th state of the latent variable, for $i=1, \ldots, K$. At the whole population level the data will then be distributed according to the mixture $p({\bf x}) = \sum_{i=1}^K P(\xi_i)p({\bf x}|\xi_i)$. Attempts to apply a GMM to model $p({\bf x})$ will not necessarily result in a one-to-one relationship between the Gaussian components in the GMM and the state-specific generative models $p({\bf x}|\xi_i)$. In general, at the very least, more than one Gaussian component will be needed to model $p({\bf x}|\xi_i)$. Although mixtures of mixture models offer increased modeling capabilities over plain mixture models to this end, they turned out to be unpopular due to the difficulties of estimation of their parameters \cite{ASteBFT15} and their excessive sensitivity to local maxima of their criterion function (generally the likelihood of their parameters given the data).

\subsection{The proposed DNN-based model}

Aiming at overcoming the aforementioned difficulties encountered with the established approaches, and due to the unexploited potential of using mixture models, the paper proposes a plausible solution in the form of a mixture model built on deep neural networks. The model, called deep neural mixture model (DNMM), relies on a convex combination of component pdfs estimated by component-specific DNNs. The DNMM belongs to the broad family of  non-parametric pdf estimation techniques, insofar that (differently form, say, GMMs) no assumptions on the form of the individual component pdfs are made \cite{DudHar73}. Due to the learning and generalization capabilities of DNNs, the DNMM can actually learn a general form for the mixture at hand, overcoming the drawbacks of the traditional non-parametric techniques, as well. A ML training algorithm is devised, satisfying (at least numerically) a combination of hard and soft constraints required in order to guarantee a proper probabilistic interpretation of the estimated model. The resulting machine can also be seen as a novel, special case of mixture of experts \cite{Yuksel2012TwentyYO} having a specific task, a ML-based unsupervised training algorithm, and a particular probabilistic strategy for assigning credit to its individual experts.

This paper is the extended, journal version of a previous workshop communication that we presented in \cite{Trentin-NMM-ANNPR18}. Over that communication, the present article extends the treatment to DNNs, defines formally the modeling capabilities of DNMM, presents the cross-validated likelihood algorithm for DNMM model selection, and reports on the results of experiments conducted over complex, multi-dimensional pdfs (in \cite{Trentin-NMM-ANNPR18} only experiments on univariate setups are reported and analyzed).

\subsection{Overview of the paper}

The paper is organized as follows. Section \ref{sec:algorithm} is devoted to the ML-based learning algorithm for DNMMs. The family of pdfs that can actually be estimated to any degree of precision via DNMM is defined formally in Section \ref{sec:theory}. An automatic model selection algorithm that exploits the probabilistic nature of the DNMM along with the cross-validated likelihood criterion is presented in Section \ref{sec:model-selection}.  The  experimental evaluation is reported in Section \ref{sec:exp}, where the DNMM compares favorably with respect to established statistical techniques (parametric as well as non-parametric) in the task of estimating non-trivial mixtures densities having different number of components and diverse dimensionality of their definition domain. Finally, Section \ref{sec:conclusions} draws the conclusions and outlines the current research directions.

\section{DNMM: formal definition and training algorithm}
\label{sec:algorithm}
\sloppy

Hereafter we rely on the notation introduced at the beginning of Section \ref{sec:intro}, such that our goal is to estimate $p({\bf x})$ from the information encapsulated within the training sample ${\cal T}$. To this aim, we introduce a deep neural mixture model ${\tilde p}({\bf x}|W)$ having the following form:

\begin{equation}
{\tilde p}({\bf x}|W) = \sum_{i=1}^{K} c_i  {\tilde p}_i({\bf x}|W_i)
  \label{eq:NMM}
\end{equation}

\noindent where $W$ represents the whole set of parameters in the DNMM (that is $c_1, \ldots, c_K, W_1, \ldots, W_k$). The mixing parameters $c_i$ shall satisfy $c_i \in [0,1]$ for $i=1, \ldots, K$ and $\sum_{i=1}^{K} c_i =1$. The $i$-th component pdf ${\tilde p}_i({\bf x}|W_i)$ is defined, for $i=1, \ldots, K$, as

\begin{equation}
{\tilde p}_i({\bf x}|W_i) = \frac{\varphi_i({\bf x}, W_i)}{\int \varphi_i({\bf x}, W_i) d{\bf x}}
  \label{eq:component}
\end{equation}

\noindent where $\varphi_i({\bf x}, W_i)$ is the function computed by a component-specific DNN whose set of learnable parameters is $W_i$. We say that this DNN realizes the $i$-th deep neural component of the DNMM. Henceforth, plain feed-forward DNNs are assumed (with arbitrary activation functions), but the following calculations can be adapted to a variety of alternate families of DNNs in a (mostly) straightforward manner. In order for the DNMM to satisfy Kolmogorov's axioms of probability, a constraint on $\int \varphi_i({\bf x}, W_i) d{\bf x}$ shall be imposed shortly. It goes without saying that each  DNN in the DNMM has $d$ input neurons (matching the dimensionality of the feature space) and a single output neuron (yielding the value of the estimated pdf), and it is expected to have as many hidden layers as needed (the automatic model selection procedure presented in Section \ref{sec:model-selection} is suitable for architecture selection, as well). Without loss of generality for all the present intents and purposes, we assume\footnote{It is seen that, in practical applications, any data normalization approach may be applied to the training sample in order to ensure respect of this assumption.}  that the random vectors of interest lie within a compact $S \subset \mathbb{R}^d$, such that $S$ can be regarded as the definition domain of $\varphi_i({\bf x}, W_i)$ for all $i=1, \ldots, K$. In so doing, numerical integration algorithms are viable to compute $\int \varphi_i({\bf x}, W_i) d{\bf x}$, as well as the other integrals required shortly. Equation (\ref{eq:component}) reduces to ${\tilde p}_i({\bf x}|W_i) = \frac{\varphi_i({\bf x}, W_i)}{\int_S \varphi_i({\bf x}, W_i) d{\bf x}}$.

The choice of the form of the activation function $f_i(.)$ used in the output layer of the $i$-th DNN requires some precautions. Due to the very nature of pdfs, $f_i(.)$ is expected to have (at least in principle) a codomain defined as $[0, +\infty)$. This may be granted in several different ways. In this paper we use a logistic sigmoid with component-specific adaptive amplitude $\lambda_i \in \mathbb{R}^+$, that is $f_i(a_i) = \lambda_i/(1+\exp(-a_i))$ as described in \cite{Trentin2001}, where $a_i$ represents the current activation value for the output neuron of the $i$-th DNN in the mixture. Therefore, each neural component in the DNMM can stretch its output over any (component-specific) range $[0, \lambda_i)$, which, instead of being predefined by the user,  is learned from the data (just like any other parameter in $W_i$) in compliance with the nature of the corresponding component pdf\footnote{Other advantages entailed by the use of adaptive amplitudes are pointed out in \cite{Trentin2001}.}.
    
The DNMM training algorithm revolves around a learning rule for the mixture parameters $W$ given ${\cal T}$, such that at the end of the training process the quantity ${\tilde p}({\bf x}| W)$ results in a robust estimate of $p({\bf x})$. This is achieved by pursuing two purposes: (1) exploiting the information encapsulated in ${\cal T}$ to approximate the unknown pdf; (2) preventing the DNNs in the DNMM from developing spurious solutions, by enforcing the constraints $\int_S \varphi_i({\bf x}, W_i) d{\bf x} = 1$ for all $i=1, \ldots, K$. To this aim, a constrained algorithm is devised that builds on the stochastic gradient-ascent maximization of the point-wise likelihood ${\tilde p}({\bf x}_j | W)$ of the DNMM given the current training pattern ${\bf x}_j$. The stochastic optimization step has to be applied iteratively for $j=1, \ldots, n$. This is achieved by means of an on-line, differentiable loss function $C(.)$ defined as

\begin{equation}
  C(W,{\bf x}_j) = {\tilde p}({\bf x}_j| W)  - \rho \sum_{i=1}^{K} \frac{1}{2} \left ( 1 - \int_S \varphi_i({\bf x}, W_i) d{\bf x} \right )^2 
\label{eq:criterion}
\end{equation}

\noindent to be maximized with respect to the DNMM parameters $W$ under the (hard) constraints that $c_i \in [0,1]$ for $i=1, \ldots, K$ and $\sum_{i=1}^{K} c_i =1$. The second term in the loss function is rather a ``soft'' constraint that enforces a unit
integral of ${\tilde p}_i({\bf x}, W_i)$ over $S$ for all $i=1, \ldots, K$, as sought, such that $\int_S {\tilde p}({\bf x} | W) d {\bf x} \simeq 1$. The hyper-parameter
$\rho \in \mathbb{R}^+$ balances the importance of the constraints, and
may be used in real-life applications in order to tackle potential numerical issues. The learning rule $\Delta w$ for a generic parameter $w$ in the DNMM is written as 
$\Delta w = \eta \frac{\partial C(.)}{\partial w}$, where $\eta \in \mathbb{R}^+$ is the learning rate.  Different calculations are needed, depending on $w$ being either (i) a mixing coefficient, i.e. $w = c_k$, or (ii) a parameter\footnote{A connection weight, bias, adaptive amplitude, or any other trainable parameter.} within any of the DNN-based component densities. In case (i), we introduce $K$ unconstrained latent  variables $\gamma_1, \dots, \gamma_K$, and we let

%\small
\begin{equation}
c_k = \frac{\varsigma(\gamma_k)}{\sum_{i=1}^{K}\varsigma(\gamma_i)}
\label{eq:mixing-par}
\end{equation}
%\normalsize

\noindent for $k=1,\ldots,K$, where $\varsigma(x) = 1/(1+e^{-x})$. In so doing, hereafter each $\gamma_k$ can be treated as the unknown parameter to be estimated, in place of the corresponding $c_k$. Consequently, higher-likelihood mixing parameters $c_k$ that satisfy the required constraints are implicitly yielded by the application of the learning rule. The latter can be written as:

\begin{eqnarray}
  \Delta \gamma_k & = & \eta \frac{\partial C(.)}{\partial \gamma_k}\\  \nonumber
  & = & \eta \frac{\partial {\tilde p}({\bf x}_j| W)}{\partial \gamma_k}\\  \nonumber
  & = & \eta \frac{\partial \sum_{i=1}^{K} c_i  {\tilde p}_i({\bf x}_j|W_i)}{\partial \gamma_k}\\  \nonumber
%  & = & \eta \sum_{i=1}^{K} \frac{\partial {\tilde p}({\bf x}_j| W)}{\partial c_i} \frac{\partial c_i}{\partial \gamma_k}\\ \nonumber
& = & \eta \sum_{i=1}^{K} {\tilde p}_i({\bf x}_j| W_i)  \frac{\partial}{\partial \gamma_k}\left(\frac{\varsigma(\gamma_i)}{\sum_{\ell=1}^{K}\varsigma(\gamma_\ell)}\right)  \\ \nonumber
& = &  \eta \left\{ {\tilde p}_k({\bf x}_j| W_k) \frac{\varsigma^\prime(\gamma_k)}{\sum_{\ell=1}^{K} \varsigma(\gamma_\ell)} - \sum_{i=1}^{K} {\tilde p}_i({\bf x}_j| W_i) \frac{\varsigma(\gamma_i)\varsigma^\prime(\gamma_k)}{[ \sum_\ell \varsigma(\gamma_\ell)]^2} \right\}  \\ \nonumber
& = & \eta \frac{\varsigma^\prime(\gamma_k)}{\sum_\ell \varsigma(\gamma_\ell)}
\left\{ {\tilde p}_k({\bf x}_j| W_k) -  {\tilde p}({\bf x}_j| W) \right\}
%\normalsize
\label{eq:delta-mixing-par}
\end{eqnarray}

Next, let us focus on scenario (ii), i.e. let $w$ be a parameter within one of the DNNs. Taking the partial derivative of $C(W,{{\bf x}_j})$ with respect to $w$ requires calculating the derivatives of the first and the second terms in the right-hand side of Equation (\ref{eq:criterion}), respectively. For notational convenience, hereafter we assume that $w$ belongs to the generic $k$-th DNN. The partial derivative of the first term can be written as

\small
\begin{eqnarray}
  \label{eq:partialC}
  \frac{\partial {\tilde p}({\bf x_j}| W)}{\partial w}  & = & \frac{\partial}{\partial w} \sum_{i=1}^{K} c_i {\tilde p}_i({\bf x_j} | W_i)   \\ \nonumber
  & = & \frac{\partial}{\partial w} \{c_k {\tilde p}_k({\bf x_j} | W_k)\}   \\ \nonumber
  & = &  c_k \frac{\partial}{\partial w} \left\{ \frac{\varphi_k({\bf x}_j, W_k)}{\int_S \varphi_k({\bf x}, W_k)d{\bf x}} \right\} \\ \nonumber
  & = & c_k \left\{ \frac{1}{\int_S \varphi_k({\bf x}, W_k)d{\bf x}} \frac{\partial \varphi_k({\bf x}_j, W_k)}{\partial w} - \frac{{\tilde p}_k({\bf x}_j, W_k)}{\int_S \varphi_k({\bf x}, W_k)d{\bf x}} \frac{\partial}{\partial w} \int_S \varphi_k({\bf x}, W_k)d{\bf x} \right\}  \\ \nonumber
  & = & \frac{c_k}{\int_S \varphi_k({\bf x}, W_k)d{\bf x}} \left\{  \frac{\partial \varphi_k({\bf x}_j, W_k)}{\partial w} - \frac{ \varphi_k({\bf x}_j, W_k)}{\int_S \varphi_k({\bf x}, W_k)d{\bf x}} \int_S \frac{\partial \varphi_k({\bf x}, W_k)}{\partial w} d{\bf x} \right\}
\end{eqnarray}
\normalsize

\noindent where we used Leibniz rule in the last step of the
calculations. In passing, it is worth noting that Equation (\ref{eq:partialC}) is the mathematical statement of the very rationale behind the specific impact that the current training pattern ${\bf x}_j$ has on the learning process for distinct neural components of the DNMM. In fact, the amount of parameter change $\Delta w$ is proportional to the probabilistic ``credit'' $c_k$ of the neural component at hand.  Moreover, the quantities within brackets in Equation (\ref{eq:partialC}) depend on the value of the $k$-th DNN output over ${\bf x}_j$, as well as on its derivative. If, at any given time during the training process, $\varphi_k(.)$ does not change significantly in a neighborhood of ${\bf x}_j$ (e.g. if ${\bf x}_j$ lies in a high-likelihood plateau or, vice versa, in a close-to-zero plateau of $\varphi_k(.)$) then the contribution of the first quantity within brackets is neglectable. Moreover, if $\varphi_k({\bf x}_j) \simeq 0$ then the second term within brackets turns out to be neglectable, as well.  To the contrary, the contribution of ${\bf x}_j$ to the parameter adaptation of $k$-th component DNN turns out to be paramount if $\varphi_k(.)$ returns a high likelihood over ${\bf x}_j$ and significant variations in its surroundings. 

Next, Leibniz rule is used again in the calculation of the derivative
of the second term in the right-hand side of Equation (\ref{eq:criterion}), which can be written as 

\begin{eqnarray}
\label{eq:constraint}
\frac{\partial}{\partial w} \left\{  \rho \sum_{i=1}^{K} \frac{1}{2} \left ( 1 - \int_S \varphi_i({\bf x}, W_i) d{\bf x} \right )^2 \right\} =  \\ \nonumber
= \frac{\partial}{\partial w}  \left\{ \frac{\rho}{2} \left ( 1 - \int_S \varphi_k({\bf x}, W_k) d{\bf x} \right )^2 \right\} =  \\ \nonumber
= - \rho \left ( 1 - \int_S \varphi_k({\bf x}, W_k) d{\bf x} \right ) \frac{\partial}{\partial w} \int_S \varphi_k({\bf x}, W_k) d{\bf x}   \\ \nonumber
= - \rho \left ( 1 - \int_S \varphi_k({\bf x}, W_k) d{\bf x} \right )  \int_S  \frac{\partial \varphi_k({\bf x}, W_k)} {\partial w} d{\bf x}.
\end{eqnarray}

\noindent In order to compute the right-hand side of Equations (\ref{eq:partialC}) and (\ref{eq:constraint}) we need suitable algorithms for the computation of $ \frac{\partial \varphi_k({\bf x}_j, W_k)}{\partial w}$, $\int_S \varphi_k({\bf x}, W_k)d{\bf x}$, and $ \int_S  \frac{\partial}{\partial w} \varphi_k({\bf x}, W_k)d{\bf x}$. As regards the computation of $ \frac{\partial \varphi_k({\bf x}_j, W_k)}{\partial w}$ we proceed as in traditional BP, or as in \cite{Trentin2001} in case $w = \lambda_k$. As for the integrals, deterministic numerical quadrature integration techniques (e.g., Simpson's method, trapezoidal rule, etc.) are viable only if $d =1$. In fact, in terms of computational burden, they cannot realistically scale up to higher dimensions ($d \geq 2$). This is all the more critical in the light of the fact that  $ \int_S  \frac{\partial}{\partial w} \varphi_k({\bf x}, W_k)d{\bf x}$ shall be iteratively computed for each parameter of each DNN in the DNMM. Moreover, deterministic integration methods do not exploit at all the nature of the function under integration. Roughly speaking, herein the integrand is expected to be an ``approximation'' of  the pdf (say, $p_k({\bf x})$) that explains the distribution of that specific sub-sample of ${\cal T}$ that is drawn from the $k$-th component of the mixture. To the contrary, accounting for the pdf of the data should drive the integration algorithm towards integration points that cover ``interesting'' regions\footnote{That is, regions having high component-specific likelihood.} of the domain of the integrand. For these reasons, we apply a component-oriented version of the approach we presented in \cite{Trentin_Soft_NEPL}. The resulting approach can be seen as an instance of Markov chain Monte Carlo \cite{MCMC-03}. It is a non-deterministic, multi-dimensional integration technique which accounts for the component-specific probability distribution of the data. Let $\phi_k({\bf x})$ denote the integrand at hand (i.e., $\varphi_k({\bf x}, W_k)$ or $\frac{\partial \varphi_k({\bf x}, W_k)}{\partial w}$). Monte Carlo with importance sampling \cite{ImportanceSampling} yields an approximation of the integral of  $\phi_k({\bf x})$ over $S$ in the form $\int_S \phi_k({\bf x}) d{\bf x} \simeq \frac{V(S)}{m} \sum_{\ell=1}^m  \phi_k(\dot{{\bf x}}_\ell)$  where $m$ properly sampled integration points $\dot{{\bf x}}_1, \ldots, \dot{{\bf x}}_m$ are used. Sampling of the $\ell$-th integration point $\dot{{\bf x}}_\ell$ (for $\ell =1, \ldots, m$) is obtained by drawing it at random from the mixture pdf $p^{(k)}_u({\bf x})$ defined as

\begin{equation}
  p_u^{(k)}({\bf x}) = \alpha(t) u({\bf x}) + (1 - \alpha(t)){\tilde p}_k({\bf x}|W_k)
  \label{eq:mixture}
\end{equation}

\noindent where $u({\bf x})$ is the uniform distribution over $S$, and $\alpha:\mathbb{N} \rightarrow (0,1)$ is a decaying function of the number $t$ of the DNMM training epochs\footnote{In the present context, a training epoch is a completed re-iteration of Equations (\ref{eq:partialC}) and (\ref{eq:constraint}) for all the parameters of the DNMM over all the observations in ${\cal T}$.} for $t=1, \ldots, T$, such that $\alpha(1) \simeq 1.0$ and $\alpha(T) \simeq 0.0$. As in \cite{Trentin_Soft_NEPL} we let  $\alpha(t) = 1/ (1 + e^{\frac{t/T - 1/2}{\theta} })$, where $\theta$ is a hyperparameter. Equation (\ref{eq:mixture})  is such that the importance sampling mechanism it implies  accounts for the (estimated) component density ${\tilde p}_k({\bf x} | W_k)$ of the $k$-th latent subsample of the data, therefore respecting the natural distribution of such sub-population and focusing integration on the relevant integration points (i.e., the points having high component-specific likelihood).  At the same time, since the estimates of this component pdfs are unfit during the early stage of the DNMM training, Equation (\ref{eq:mixture}) prescribes a (safer) sampling from a uniform distribution at the beginning (practically behaving like a plain Monte Carlo algorithm). As long as the robustness of the DNMM estimate increases, i.e. as $t$ increases, sampling from  ${\tilde p}_k({\bf x} | W_k)$ replaces progressively the sampling from $u({\bf x})$, ending up in purely non-uniform importance sampling. The form of $\alpha(t)$ is defined accordingly. Since $\varphi_k({\bf x}, W_k)$ is intrinsically non-negative, for $t \rightarrow T$ the sampling occurs substantially from $\left\vert{\varphi_k({\bf x}, W_k)}\right\vert /\int_S \left\vert{\varphi_k({\bf x}, W_k)}\right\vert d{\bf x}$, that is a sufficient condition for granting that the variance of the estimated integral vanishes and the corresponding error goes to zero \cite{Ohl:1998jn}. 

Sampling  from $p^{(k)}_u({\bf x})$ requires a viable technique for sampling from the $k$-th DNN in the DNMM. A variant of Markov chain Monte Carlo, namely the Metropolis--Hastings (M-H) algorithm \cite{Newman1999book}, is exploited in this paper. M-H is robust to the fact that, during training, $\varphi_k({\bf x}, W_k)$ may not be properly normalized but it is proportional by construction to the corresponding pdf estimate (which is normalized properly, instead, by definition) \cite{Newman1999book}. Due to its efficiency and ease of
sampling, a multivariate logistic pdf with radial basis, having location ${\bf x}$ and scale $\sigma$, is used as the {\em proposal} pdf $q({\bf x}^\prime | {\bf x})$ required by M-H to generate a new candidate sample ${\bf x}^\prime = (x^\prime_1, \ldots, x^\prime_d)$ from the current sample ${\bf x} = (x_1, \ldots, x_d)$. Formally, such a proposal pdf is defined as $q({\bf x}^\prime | {\bf x}, \sigma) = \prod_{i=1}^d \frac{1}{\sigma} e^{(x^\prime_i - x_i)/\sigma} (1 + e^{(x^\prime_i -x_i)/\sigma})^{-2}$ which can be easily sampled via the inverse transform sampling technique. The hyperparameters needed (i.e., the scale $\sigma$ of the proposal pdf and the burn-in period for M-H) are fixed empirically as part of the overall model selection process (see Section \ref{sec:model-selection}).

\section{Class of pdfs that can be modeled accurately via DNMMs}
\label{sec:theory}

It is seen that not all pdfs can be estimated in an accurate manner via DNMMs. A couple of simple, univariate examples should be more than enough to convince the reader of this: the Dirac's Delta (which is not continuous and not bounded) and the standard exponential pdf (defined as $p_e(x) =0$ if $x<0$, $p_e(x) = \exp(-x)$ otherwise). In \cite{TrentinLC18} we introduced the class of {\em nonpaltry pdfs}, that basically embrace all the ``interesting'' pdfs that are of practical interest (and, that can  be estimated to any degree of precision by means of Parzen Neural Networks). The arguments handed out by \cite{TrentinLC18} can be extended to DNMMs, as well. The formal definition of this class of function goes as follows. Let $S$ be a compact subset of $I_d$, where $I_d = [0,1]^d$ (the symbol $S$ has the same practical meaning it had in the previous section). A continuous pdf $p:\mathbb{R}^d \rightarrow \mathbb{R}$ is said to be {\em nonpaltry} over $S$ if  $\int_{S} p({\bf x}) d {\bf x} = 1$. Accordingly, we formally define the class ${\cal P}(S)$ of nonpaltry pdfs over $S$ as the set of all the density functions that are nonpaltry over $S$. In passing, note that mixture densities built on nonpaltry component densities are, in turn, nonpaltry. Theorem 3.1 in \cite{TrentinLC18} states that for any nonpaltry pdf $p(\cdot)$ over $S$ at least one DNN exists that computes $\phi^{(\epsilon)}(\cdot)$ s.t. $\| \phi^{(\epsilon)}(\cdot) - p(\cdot) \|_{\infty,S} < \epsilon$ for any $\epsilon \in \mathbb{R}^+$. This (trivially) implies that at least one DNMM (with $c$ =1) exists that computes $\phi_{DNMM}^{(\epsilon)}(\cdot)$ s.t. $\| \phi_{DNMM}^{(\epsilon)}(\cdot) - p(\cdot) \|_{\infty,S} < \epsilon$ for any $\epsilon \in \mathbb{R}^+$. Roughly speaking, we can estimate any nonpaltry pdf to any degree of precision by means of a DNMM.

\section{Automatic likelihood-driven model selection strategy}
\label{sec:model-selection}

The former section presented a modeling (or, approximation) capability of the family of pdfs realized by DNMMs, regardless of the actual convergence (during training) of a specific DNMM to the unknown pdf $p(\cdot)$ underlying the data ${\cal T}$ at hand. In practice, roughly speaking, the larger the cardinality of the training sample ${\cal T}$, the closer to $p(\cdot)$ it tends to become (in probability) the estimate realized by the DNMM. This means that, for sufficiently large values of $n$, the likelihood of the DNMM given a separate validation set ${\cal V}$ (independently drawn from $p(\cdot)$, as well) tends to ``increase\footnote{The term ``increase'' is used hereby with its qualitative meaning, that is to indicate a trend and not a strictly monotonic behavior: in fact, the likelihood of the DNMM does not necessarily increase monotonically as the training process proceeds.}'' with $n$. This informal reasoning suggests the adoption of the maximum-likelihood criterion as an evaluation function that can be used throughout the DNMM model selection process. This is made viable by the very probabilistic nature of the DNMM, and it is not available to generic, non-probabilistic DNNs or mixtures of experts.

In short, an iterative model/hyperparameter search strategy may be developed as follows.To fix ideas, let us first consider the issue of fixing the number of neurons in a certain hidden layer of a DNN within the DNMM\footnote{Note that there is no assumption in the definition of DNMM that forces the practitioner to use the same number of layers and/or neurons in the different DNNs realizing the components of the mixture, although this may be a computationally realistic choice.}. We start with an initial number of hidden neurons, and we increase it by $u$ units (for a certain integer $u$, e.g. $u = 1$) at each of the following steps of the selection procedure. At each iteration (starting from the initialization) the DNMM is trained on ${\cal T}$, and the likelihood of the resulting model given ${\cal V}$ is evaluated. The procedure is iterated for as long as this likelihood increases. We stop searching when, for $\tau$ consecutive iterations, the relative gain in likelihood between consecutive steps is lower than a fixed percentage $\nu \%$. The minimum description length principle is then applied \cite{MDL-J1978465}, selecting the simplest model amongst those (explored throughout the last $\tau$ iterations) which yielded comparable values of the corresponding validation likelihood given ${\cal V}$ (where ``comparable'' means that their difference is below a user-defined threshold). It is worth noting that this approach relies on an instance of the so-called cross-validated likelihood criterion \cite{Rust1985}. The same incremental strategy is viable for selecting the number of layers within the DNNs in the mixture (e.g., starting from a single hidden layer and deepening the architecture by adding one more layer at each step).

The present cross-validated likelihood search strategy can be applied, in a straightforward manner, to the problem of selecting the other hyperparameters controlling the behavior of the DNMM learning process. In order to limit the computational burden of such an automatic model selection process, random search of the values of the hyperparameters (each choice characterized by the corresponding likelihood given ${\cal V}$) is recommended instead of going through the step-by-step incremental approach we proposed for selecting the umber of neurons.

\section{Experiments}
\label{sec:exp}
\sloppy

Experiments are conducted on random samples drawn from multimodal mixtures having known form. Section \ref{sec:univariate} reports on the results obtained on univariate data using mixtures of 3-layer DNNs, while Section \ref{sec:multivariate} presents the results of experiments involving multivariate data (of different dimensionalities) using deeper DNMMs.

\subsection{Experiments with univariate data}
\label{sec:univariate}

In the present, univariate case, the random samples were drawn from mixtures $p(x)$ of $c$ Fisher-Tippett pdfs, i.e. $p(x) = \sum_{i=1}^{c}\frac{P_i}{\beta_i} \exp \left(-\frac{x-\mu_i}{\beta_i}\right) \exp \left\{ -\exp \left(-\frac{x-\mu_i}{\beta_i}\right) \right\}$. The mixing coefficients $P_1, \ldots, P_c$ were drawn at random from the uniform distribution over $[0.1]$ and normalized in such a way that $\sum_{i=1}^c  P_i = 1$.  The component densities of the Fisher-Tippett mixture are identified by their scales $\beta_i$ and locations $\mu_i$, for $i=1, \ldots, c$. The scales were drawn at random from the uniform distribution over $(0.01, 0.9)$, and the locations were randomly and uniformly distributed over $(0,10)$. Estimation tasks were created using $1200$ random patterns each, drawn from $p(x)$, and a variable number $c$ of component densities (namely $c=5$, $10$, $15$ and $20$). Each $c$-specific sample was split into a training set (with $n = 800$ patterns) and a validation set (the remaining $400$ patterns).  The integrated squared error (ISE) between $p(x)$ and its estimate ${\tilde p}(x)$, i.e. $\int \left(p(x) - {\tilde p}(x)\right)^2 d x$, is adopted in order to assess the performance of the different estimators. Simpson's method was applied to compute numerically the ISE.

In the present, illustrative setup we used DNMMs involving DNNs with a 3-layer architecture (input layer, hidden layer of  9 neurons, output layer). Sigmoid activation functions were used in the hidden and the output layers, having layer-wise  \cite{Trentin2001} adaptive $\lambda$. All the DNMM parameters underwent random initialization over zero-centered intervals, except for the values of $\lambda$ (that were initialized to $1$) and the mixing parameters that were initialized as $c_i = 1/K$ for $i=1, \ldots, K$. As in \cite{Trentin_Soft_NEPL} we used a function $\alpha(t)$ having $\theta = 0.07$, and we relied on $m=400$ integration point. The latter ones were sampled at the beginning of each training epoch using a scale $\sigma = 9$ for the logistic proposal pdf in M-H. The burn-in period of the Markov chain in M-H was stretched over the first $500$ states. The other hyperparameters of the DNMM were fixed via random-search based on the cross-validated likelihood criterion presented in Section \ref{sec:model-selection}. No normalization was applied to the values of the input data. Table \ref{tab:comp} reports the results. DNMMs having different values of $K$ (that is $K=4$, $8$ and $12$) were tested and compared with 8-GMM, 16-GMM, and 32-GMM (initialized via $k$-means and refined via iterative ML \cite{DudHar73}), $k_n$-NN with unbiased $k_n = 1 \sqrt{n}$ \cite{DudHar73}, and Parzen Window (PW) with standard width $h_n = 1/\sqrt{n}$ of its Gaussian kernels \cite{DudHar73}. 

\begin{table}[htp]
%\centering
\caption{Estimation of the Fisher-Tippett  mixture $p(x)$ (with $n=800$) in terms of integrated squared error as a function of the number $c$ of the Fisher-Tippett  component densities. Best results are shown in boldface. ({\em Legend}: $k$-GMM and $k$-DNMM denote the GMM and the DNMM with $k$ components, respectively).}
\label{tab:comp}
%      {\normalsize
\vspace{0.2cm}
\begin{small}       
\begin{tabular}{l l @{\hskip 0.4cm} l @{\hskip 0.4cm} l @{\hskip 0.4cm} l|l}
\hline
$c$  & 5 & 10 & 15 & 20 & {\em Avg. $\pm$ std. dev.}\\
\hline
8-GMM & $9.60\textrm{e}-3$ & $1.12\textrm{e}-2$ & $4.57\textrm{e}-2$ & $7.99\textrm{e}-2$ &  $(3.66 \pm 2.89) \textrm{e}-2$ \\
16-GMM & $6.33\textrm{e}-3$ & $9.29\textrm{e}-3$ & $3.78\textrm{e}-2$ & $4.24\textrm{e}-2$ & $(2.40 \pm 1.63) \textrm{e}-2$  \\
32-GMM & $7.15\textrm{e}-3$ & $9.82\textrm{e}-3$ & $2.41\textrm{e}-2$ & $3.03\textrm{e}-2$ & $(1.78 \pm 0.97) \textrm{e}-2$  \\
$k_n$-NN & $6.54\textrm{e}-3$ & $8.70\textrm{e}-3$ & $2.03\textrm{e}-2$ & $2.36\textrm{e}-2$ & $(1.48 \pm 0.73)\textrm{e}-2$  \\
PW & $6.02\textrm{e}-3$ & $8.94\textrm{e}-3$ & $2.14\textrm{e}-2$ & 1$.98\textrm{e}-2$ & $(1.40 \pm 0.67)\textrm{e}-2$  \\
4-DNMM & $6.41\textrm{e}-3$ & $7.06\textrm{e}-3$ & $1.09\textrm{e}-2$ & $1.40\textrm{e}-2$ & $(9.59 \pm 3.07) \textrm{e}-3$  \\
8-DNMM & ${\bf 5.89\textrm{e}-3}$ & ${\bf 6.02\textrm{e}-3}$ & $8.11\textrm{e}-3$ & $1.01\textrm{e}-2$ & $({\bf 7.53} \pm 1.73) \textrm{e}-3$ \\
12-DNMM & $6.38\textrm{e}-3$ & $6.27\textrm{e}-3$ & ${\bf 8.05\textrm{e}-3}$ & ${\bf 9.64\textrm{e}-3}$ & $(7.59  \pm {\bf 1.38}) \textrm{e}-3$
\end{tabular}
\end{small}
%}
\end{table}

The Table shows that, regardless of the model used, the corresponding ISE increase as a function of $c$, as expected. The DNMMs improve systematically (and, significantly) over the statistical techniques, whatever the value of of $c$. On average, both the 8-DNMM and the 16-DNMM yield a $46\%$ relative ISE reduction over the PW (the latter turns out to be the most robust non-neural estimation algorithm). Welch's t-test (used in order to account for the different variances of the models) \cite{Japkowicz:2011} returns a level of confidence $> 90\%$ on the statistical significance of the gap between the 8- (or, 12-) DNMM and the PW. Furthermore, the DNMMs  turn out to be the stablest models overall, as proved by the values of the corresponding standard deviations (last column of the table). This is evidence of the fact that the estimation accuracy offered by the DNMMs is less sensitive to the complexity of the underlying Fisher-Tippett mixture (i.e., to $c$) than the accuracies yielded by the traditional statistical techniques. Finally, differences in terms of ISE are observed among the DNMMs depending on $k$. Nonetheless, differences between the 8-DNMM and the 12-DNMM turn out to be mild, and they depend on the complexity of the underlying pdf to be estimated (at least to some extent), as expected.

\subsection{Experiments with multivariate data}
\label{sec:multivariate}

Hereafter we use multivariate data drawn from mixtures of generalized extreme value distributions (GEV) \cite{castillo2004extreme} with null slope and having the following parametric form:

\begin{equation*}
{\text{\em m-GEV}}({\bf x};{\bf \Theta} ) = \sum_{k=1}^{c_{T}} \frac{1}{c_{T}} \prod_{i=1}^{d} \frac{1}{\beta_{ki}} \exp
\left(-\frac{x_i -\mu_{ki}}{\beta_{ki}}\right) \exp \left\{ -\exp
\left(-\frac{x_i - \mu_{ki}}{\beta_{ki}}\right) \right\}
\label{eq:GEV}
\end{equation*}

\noindent where the GEV parameters ${\bf \Theta} = (d, c_{T}, {\bm \mu}_1, \ldots, {\bm \mu}_{c_T}, {\bm \beta}_1, \ldots, {\bm \beta}_{c_T})$ are defined as follows:

\begin{enumerate}
  
\item $d$ is the dimensionality of the feature space as usual. Therefore, the generic real valued random vector ${\bf x} \in \mathbb{R}^d$ can be written as ${\bf x} = (x_1, \ldots, x_d)$;

\item $c_T$ is the number of component GEVs in the mixture, and $c_T = c^d$ where $c$ denotes the (fixed) number of diverse modes of ${\text{\em m-GEV}}({\bf x};{\bf \Theta} )$ along each one of the dimensions in the definition domain of the pdf;

\item ${\bm \mu}_k = (\mu_{k1}, \ldots, \mu_{kd})$ and ${\bm \beta}_k = (\beta_{k1}, \ldots, \beta_{kd})$ are the location and the scale vectors of $k$-th component GEV, respectively. They are defined in such a manner that the set $\{({\bm \mu}_k, {\bm \beta}_k) | k=1, \ldots, c^d\}$ embraces all possible combinations of $c$ dimension-specific parameters $({\bar \mu}_{i1}, {\bar \beta}_{i1}), \ldots, ({\bar \mu}_{ic}, {\bar \beta}_{ic})$  for $i=1, \ldots, d$ (for a total of $c_T = c^d$ combinations).

\end{enumerate}

\noindent  For each $i=1,\ldots,d$ and $j=1,\ldots,c$, the values ${\bar \mu}_{ij}$ are random quantities, independently and uniformly drawn from the interval $(0.1,0.9)$, and the iid random values ${\bar \beta}_{ij}$ are drawn from the interval $(0.03,0.05)$. In the following, numerical integration is computed over the interval $S = [0, 1.1]^d$. 

\begin{table}[htp]
%\centering
\caption{Estimation of the multivariate mixture of generalized extreme value distributions  ${\text {\em m-GEV}}({\bf x};{\bf \Theta})$ (with $n=800$) in terms of relative ($\%$) ISE reduction over the baseline, as a function of the number $C_T$ of the component densities of the ${\text {\em m-GEV}}({\bf x};{\bf \Theta})$ and of the dimensionality $d$ of the feature space.}
\label{tab:mult}
%      {\normalsize
\vspace{0.2cm}
%\begin{small}       
\begin{tabular}{l l @{\hskip 0.4cm} l @{\hskip 0.4cm} l @{\hskip 0.4cm} l|l}
\hline
 & $C_T = 4$ & $C_T = 9$ & $C_T = 16$ & $C_T = 25$ & {\em Avg.}\\
\hline
$d = 2$ & 11.31 & 10.52 & 7.38 & 9.02 & $9.56 \pm 1.50$  \\
$d = 4$ & 8.44 & -0.07 & 5.75 & 7.01 & $5.28 \pm 3.23$  \\
$d = 6$ & 4.98 & -1.64 & 5.80 & 8.13 &  $4.32 \pm 3.63$ \\
$d = 8$ & 6.20 & 7.34 & 8.63 & 8.00 & $7.54 \pm 0.90$ \\
\hline
{\em Avg.} & $7.73 \pm 2.41$ & $4.04 \pm 5.05$  & $6.89 \pm 1.20$ & $8.04 \pm 0.71$ & ${\bf 6.67 \pm 3.29}$
\end{tabular}
%\end{small}
%}
\end{table}

We generated different estimation tasks involving {\em m-GEVs} with an increasing number of components ($C_T = 4, 9, 16$ and $25$, respectively) and an increasing dimensionality of the feature space ($d = 2, 4, 6$ and $8$, respectively). For each combination of such values for $C_T$ and $d$, a sample of as many as $1200$ patterns was randomly drawn from ${\text {\em m-GEV}}({\bf x};{\bf \Theta})$. Each such data sample was split into a training set ($800$ random vectors) and a validation set (the remaining $400$ patterns), as we did before. The validation set was used to select the architectures and hyperparameters for the algorithms via ML-based random search (applying the model selection procedure presented in Section \ref{sec:model-selection}) on a $d$- and $c_T$-specific basis. This cross-validated likelihood criterion was exploited also in order to fix a statistical baseline for assessing the performance of the DNMM, as follows. First, three established and popular statistical estimation techniques suitable for multivariate, multimodal pdf estimation were evaluated, namely the PW, the $k_n$-NN, and the GMM (the latter was evaluated several times, for an increasing number of component Gaussians ranging from 4 to 32). In PW we used Gaussian kernels with bandwidth $h_n = h_1/\sqrt{n}$, where $h_1$ was selected over the $[10^{-1},10^1]$ range.  For the $k_n$-NN we let  $k_n = k_1 \sqrt{n}$, as usual, with $k_1$ selected via cross-validated ML. As for the GMMs, the $k$-means algorithm was used for the initialization of the parameters, and the latter were refined via iterative ML re-estimation \cite{DudHar73}. After ML selection of suitable hyperparameters, the best amongst these statistical models (in terms of likelihood on the validation set) was used for fixing the baseline (for the specific combination of $d$ and $c_T$ at hand, then the whole procedure was repeated for the other combinations). The baseline was quantified as the ISE with respect to the true {\em m-GEV} underlying the data sample for those specific values of $d$ and $c_T$. Next, a DNMM was selected (again, based on the procedure presented in Section \ref{sec:model-selection}). Selection involved fixing the architecture of the DNNs (number of layers between 3 and 6, number of neurons per layer, type of activation functions) and the hyperparameters used for training the DNMM. The same architectures and hyperparameters were applied for all the DNNs in the DNMM at any step of the random search process. DNMMs were initialized as in the previous Section, and the same parameters used therein for the numeric integration (burn-in period, form of $alpha(t)$, $m$, etc.) were used herein. Eventually, the ISE offered by the best after-training DNMM selected this way was evaluated w.r.t. the true {\em m-GEV}, and compared with the baseline. In so doing, the results (reported in table \ref{tab:mult}) can be expressed in terms of the percentage of relative ISE reduction ($\%$) yielded by the best DNMM over the baseline\footnote{To fix ideas, a $0\%$ relative ISE reduction means that the DNMM equals the baseline, a negative relative ISE reduction means that the DNMM performance worsened the baseline, while positive reductions represent gains over the baseline (the larger, the better: ideally, in the limit case, a $100\%$ relative ISE reduction would be observed if the DNMM could result in a perfect model of the true {\em m-GEV} in presence of an imperfect estimation baseline).}. Expressing the results of the experiments in this way has two significant advantages: (1) keeping the amount of values to be read and compared, and the size of the table, to a human-readable scale, and (2) conveying the sense of the gap between the DNMM and any other established statistical estimation technique. 

It is seen that, in a quasi-systematic manner, the DNMM improves significantly over the statistical approaches. Only in two cases out of 16 the ISE yielded by the DNMM is just in line with (or, slightly worse than) the baseline. Overall, the DNMM offers an average $6.67 \pm 3.29$ relative ISE reduction over its closest competitor (that is, the baseline). The statistical significance of the improvement offered by the DNMM over the baseline (and, in turn, over any statistical technique) computed via Welch's t-test is high, namely $\geq 90\%$. Moreover, the values of the std. deviation (last column and last row of the table) show that the DNMM results in a substantially stable behavior overall.

We cannot help but pointing out that the gap between the DNMM and the best statistical estimates is not as wide as in the univariate setup. A possible rationale behind the reduced gap is likely to be found in the model selection procedure we applied: the random-search strategy applied within the cross-validated ML model selection algorithm over the multivariate data (much needed, in place of a more accurate grid search, in order to keep the overall computational burden to a realistic scale) turns out to be capable of finding suitable, yet quite sub-optimal architectures and hyperparameters. This affects significantly the model selection in the DNMM (insofar that the latter has many hyperparameters to be fixed, each choice resulting in profoundly different models and, therefore, performances), while the issue is much less severe in the case of the statistical estimators (which basically involve selecting just a few hyperparameters, even one only in the PW and the $k_n$-NN cases). Another reason for the reduced gap between the DNMM and the baseline lies in our inheriting from the univariate setup the hyperparameters for M-H: it is seen that the effectiveness of (say) a certain number $m$ of integration points (which is suitable to univariate data) reduces significantly as the dimensionality of the feature space increases (in fact, many more integration points would be needed in order to obtain a representative coverage of the multidimensional space that could grant an accurate numerical computation of the integral of the multivariate pdf). Nevertheless, also in the present setup the results are evidence of the improved estimation and modeling capabilities of the DNMM even in complex multivariate density estimation tasks.

\section{Conclusions}
\label{sec:conclusions}

The paper presented the DNMM as a sound DNN-based model for multivariate pdf estimation that satisfies Kolmogorov's axioms. The DNMM overcomes all the usual shortcomings of traditional statistical techniques. In particular, its nonparametric modeling capabilities allow for realizing flexible component densities that are not constrained to have a predefined form, as it happens in parametric statistics. Moreover, the DNMM does not suffer from the limitations of nonparametric statistical estimators, as well, insofar that it is a learning machine (instead of a memory-based algorithm), which entails generalization, smooth and compact solutions, as well as reduced time- and space-complexity. Furthermore, the probabilistic nature of the DNMM allows for the exploitation of the cross-validated likelihood criterion in order to carry out the model selection relying on the ML of the hyperparameters given the validation data.

The experimental results show significant improvements yielded by the DNMM over the baseline of the statistical techniques on both the univariate and the multivariate data drawn from several complex pdfs. Future work  revolves around the initialization procedure: non-uniform initialization of the mixing coefficients may turn out to be helpful in breaking potential ties, and initializing the individual DNN-based components via supervised learning of a subset of the components of a pre-es`<timated reference mixture model (i.e., a GMM) may improve the random initialization of the DNNs parameters.  

\bibliographystyle{spmpsci}      % mathematics and physical sciences
\bibliography{NPL} 

\end{document}